%% file: main.tex
\documentclass[10pt,twocolumn,letterpaper]{article}

\usepackage[applications]{wacv}
\usepackage{mathtools}
\usepackage{float}
\usepackage{pgfplots}

\usepackage{booktabs}
\usepackage{amsmath}
\usepackage{booktabs}
\usepackage{dblfloatfix}

\usepackage{tikz}
\usepackage{collcell}
\usepackage{booktabs}
\usepackage{rotating}
\usepackage{diagbox}
\usepackage{pgfplotstable}
\usepackage{xcolor}
\usepackage{colortbl}
\usepackage{caption}
\usepackage{graphicx}
\usepackage{subcaption}
\usepackage{cuted} % preambulumban
\usepackage{subcaption}

\usepackage{dblfloatfix}
\input{preamble}

\definecolor{wacvblue}{rgb}{0.21,0.49,0.74}
\usepackage[pagebackref,breaklinks,colorlinks,allcolors=wacvblue]{hyperref}

\title{Building temporally coherent 3D maps with VGGT for memory-efficient Semantic SLAM}

\usepackage[linesnumbered,ruled,vlined]{algorithm2e}

\author{Gergely Dinya\\
Eötvös Loránd University\\
Budapest, Hungary\\
{\tt\small aazbjw@inf.elte.hu}
\and
Péter Halász\\
Pázmány Péter Catholic University\\
Budapest, Hungary\\
{\tt\small halasz.peter@hallgato.ppke.hu}
\and
András Lőrincz\\
Eötvös Loránd University\\
Budapest, Hungary\\
{\tt\small lorincz@elte.hu}
\and
Kristóf Karacs\\
Pázmány Péter Catholic University\\
Budapest, Hungary\\
{\tt\small karacs@itk.ppke.hu}
\and
Anna Gelencsér-Horváth\\
Pázmány Péter Catholic University\\
Budapest, Hungary\\
{\tt\small gha@itk.ppke.hu}
}

\begin{document}

\twocolumn[{
\maketitle

\begin{center}
  \includegraphics[width=0.9\textwidth]{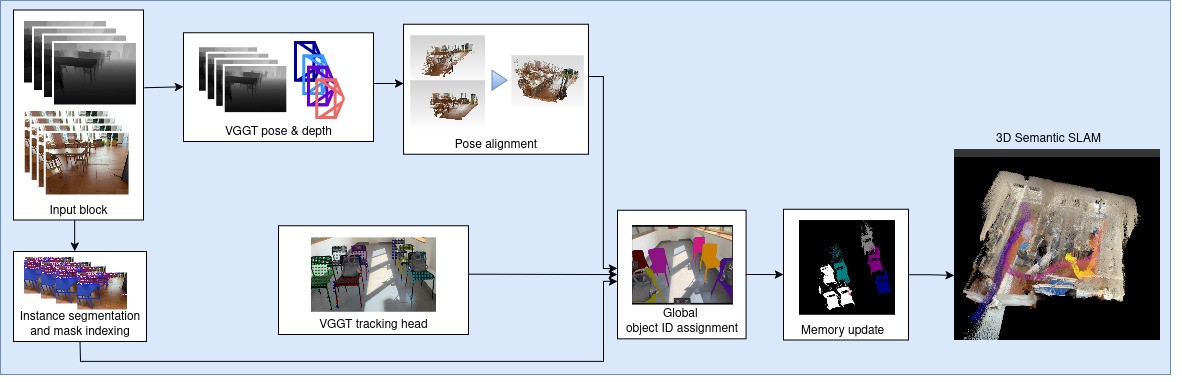}
  \captionof{figure}{Overview of the proposed pipeline: memory-efficient, temporally coherent semantic SLAM with VGGT.}
  \label{fig:pipeline}
\end{center}
}]
\input{vggt/0_abstract}

\input{vggt/1_intro_v1}
\input{vggt/2_related}

\input{vggt/3_methods}
\input{vggt/4_results}
\input{vggt/5_conclusion}

%% A double blind review miatt nem lehet benne Acknowledgment
%\input{vggt/6_ack}

%\newpage
{
    \small
    \bibliographystyle{ieeetr}
    \bibliography{main}
}

\end{document}

%% file: preamble.tex
\DeclareMathOperator*{\argmin}{argmin}

%% file: vggt/0_abstract.tex
\begin{abstract}

% We present a fast, spatio-temporal scene understanding framework 
% based on Vision Gated Generative Transformers (VGGT),
% The proposed pipeline is designed to enable an efficient, close to real-time performance, supporting applications including assistive navigation.
% Our method processes video input using a sliding window, enabling continuous updates of the 3D scene representation. We exploit the VGGT tracking head to aggregate semantic instance masks into objects and to 
% detect changes. 
% To allow for temporal consistency and richer contextual reasoning the system stores timestamps and instance-level identities. % kiegészíteni vagy törölni 
% Keyframes are selected to balance memory usage and 
% maintain track and 3D structural stability. %% 
% We evaluate the approach on well-known benchmarks and custom ones specifically designed for assistive navigation scenarios. 
% %Results show improved robustness in  environments and enhanced reliability for user guidance.
% Results show improved robustness to changes in the environment and enhanced reliability for user guidance.

We present a fast, spatio-temporal scene understanding framework 
based on Visual Geometry Grounded Transformer  (VGGT).
The proposed pipeline is designed to enable efficient, close to real-time performance, supporting applications including assistive navigation.
To achieve continuous updates of the 3D scene representation, we process the image flow with a sliding window, aligning submaps, thereby overcoming VGGT’s high memory demands. We exploit the VGGT tracking head to aggregate 2D semantic instance masks into 3D objects. To allow for temporal consistency and richer contextual reasoning the system stores timestamps and instance-level identities, thereby enabling the detection of changes in the environment. We evaluate the approach on well-known benchmarks and custom datasets specifically designed for assistive navigation scenarios. 
The results demonstrate the applicability of the framework to real-world scenarios.
\end{abstract}

%%%%% xtra lines as ideas:
% In this work, we introduce a pipeline based on Vision Gated Generative Transformers (VGGT)~\cite{vggt}. 
% Our approach processes video input through a sliding window, while a tracking head detects motion and structural changes. 
% The system stores both timestamps and instance-level identities to maintain temporal consistency. 
% Key frames are selected to balance memory usage with long-term scene stability. 

% We design the pipeline to efficiently fuse spatial and temporal cues while maintaining real-time performance. 

%% file: vggt/1_intro_v1.tex
\section{Introduction}
\label{sec:intro}

3D scene understanding underpins assistive navigation: users must operate in cluttered, unfamiliar, and evolving indoor environments. 
A practical system therefore needs a temporally coherent 3D map that stays stable across viewpoint change, partial occlusions, and genuine scene state changes. 
%
% Alongside methods like Embodied SAM~\citep{esam}, or Gaussian-splatting based approaches such as GSFusion~\citep{gsfusion} and Safer Splat~\citep{safersplat}, recent advances in transformer architectures, such as VGGT~\citep{vggt} have opened new opportunities. 
% Transformers enable strong temporal modeling and scalable representation learning. 
% This makes them a promising choice for combining spatial cues with temporal dynamics.  
%
Moreover, approaches should be designed for a low memory footprint and high speed. Feed-forward multi-view models such as VGGT~\citep{vggt} provide fast depth and pose estimation. 
However, applying this paradigm to long sequences is challenging for several reasons: (i) memory requirements grow rapidly with sequence length; (ii) temporal dependencies are not modeled explicitly; and (iii) the target application demands sustained spatio-temporal consistency across hundreds of frames while also incorporating semantics --- i.e., handling open-world objects --- and adapting to rearrangements and genuine scene changes.

In realistic deployments, input arrives as a continuous video stream and the map must be updated online. 
Methods that rely on global, post-hoc optimization over full sequences presuppose offline access to all data and are ill-suited for real-time guidance. 
Accordingly, we target an online, streaming-friendly formulation that incrementally maintains geometry and semantics as frames arrive.
\footnote{We will publish the code in a public repository.}

We focus on indoor environments, a primary deployment setting for assistive navigation, where mapping is challenged by reflective and transparent surfaces, repeated structures, motion blur from handheld sensor devices, and frequent occlusions.
We primarily target a monocular RGB setup; when auxiliary depth from commodity mobile sensors (e.g., LiDAR) is available, we use it only to stabilize metric scale, and we report results for both configurations.

Beyond integrating semantics into a geometrically accurate 3D map suitable for SLAM, the semantic layer should provide a firm basis to perform tasks with real user impact; accordingly, we consider the broadly relevant problem of finding an empty seat in indoor environments as a representative challenge.
This can be challenging in unfamiliar spaces and because seat state can change even when the surrounding scene appears static.

\noindent
Our contributions can be summarized as follows:
\begin{itemize}
    \item We introduce a memory-efficient, block-wise VGGT pipeline that enables long-sequence 3D reconstruction in a streaming setting, overcoming the prohibitive memory growth of standard feed-forward multi-view models while preserving geometric consistency.

    \item We develop a temporally coherent semantic mapping strategy that fuses 2D instance segmentation with the VGGT tracking head, producing stable, persistent 3D object identities across viewpoint changes, occlusions, and re-entries into the field of view.
    
    \item We propose a lightweight change-detection mechanism based on timestamped object trajectories and depth-based visibility reasoning, allowing the system to detect object presence, disappearance, and state changes in realistic indoor environments.
    
    \item We integrate these components into an online SLAM framework that maintains both geometry and semantics incrementally, supports real-time operation on commodity hardware, and is well suited for assistive navigation tasks such as identifying free or occupied seats.
    
    \item We provide quantitative evaluations on standard datasets and qualitative results on custom assistive-navigation scenarios, demonstrating that the method achieves competitive accuracy while remaining computationally efficient during inference.
\end{itemize}

%% file: vggt/2_related.tex
\section{Related work}
\label{sec:related}

Recent years have seen rapid progress in close to real time 3D scene representations from motion. 
We build on VGGT~\citep{vggt}, a pretrained, feed-forward vision transformer that, from RGB images alone, predicts depth, a point cloud, and camera intrinsics and poses (extrinsics).
The authors benchmark on an NVIDIA H100 GPU and report 8.75s for a 200-frame sequence. The memory footprint exceeds 40GB at 200 frames and increases approximately linearly with sequence length on this setup.

% ide tegyünk egy mondatot, hogy bár csak nyáron jött ki már sokan építenek rá amit jól mutat az alábbi, leginkább relevánsnak érzett lista?

VGGT Long~\citep{vggtLong} extends VGGT’s memory-limited, locally consistent point-map generation to kilometer-scale scenes. 
It partitions the sequence into overlapping chunks ($\sim$60-75 frames), aligns each chunk to the previous via Iteratively
Reweighted Least Squares (IRLS) optimization, and mitigates drift through loop detection and Sim(3)
%~\citep{sim3} 
alignment of overlapping pointclouds, followed by a global optimization over the chunk transforms. 
Because it requires multiple passes and access to the full sequence—including a final global adjustment over all frames—the method is not suited to streaming input; moreover, dynamic objects are typically down-weighted and effectively treated as outliers due to low confidence.

FastVGGT~\citep{fastvggt} accelerates VGGT via training-free token merging in the Global Attention module, achieving a 4$\times$ speed-up on an NVIDIA H100 GPU. 
Whereas VGGT, provided 80GB VRAM,
runs out of memory at around 300 frames,
%on the H100 GPU device, 
FastVGGT increases the feasible sequence length to over $1\,000$ frames ($\simeq 3.5\times$). 

To address true streaming/online use, StreamVGGT~\citep{streamVGGT} is designed to process incoming frames incrementally. However, because its spatio-temporal tokens scale with sequence length, both memory usage and latency grow with the number of frames.
It replaces global self-attention with temporal causal attention and caches historical keys/values as an implicit memory.
The cached-token design implies that memory footprint grows with sequence length, which the authors explicitly list as a challenge for long sequences and resource-constrained devices.

VGGT-SLAM~\citep{vggtslam} extends feed-forward VGGT to long sequences by running VGGT on overlapping keyframe windows to form submaps, then aligning them via 15-DOF homographies on the SL(4) manifold to address the projective ambiguity that arises with uncalibrated monocular, offline input.
The pipeline augments sequential alignment with loop-closure constraints and solves a global factor-graph optimization on SL(4).

Spatial Tracker V2~\citep{spatialTrackerV2} unifies 3D point tracking, monocular video depth, and camera pose estimation in a feed-forward pipeline, achieving state-of-the-art results. 
The authors report up to $50\times$ faster runtime than dynamic 3D reconstruction approaches, however, memory usage grows with sequence length, and a $\sim$150-frame input may require more than 40GB of GPU VRAM.

%3DGS~\citep{unposed3dgs}
%ORB~\citep{ORB} % features: keypts and descriptors.

%% file: vggt/3_methods.tex
\section{Methods}
\label{sec:methods}

% kéne írni arról, hogy mi a VGGT gyengesége: miért nem működik a VGGTSLAM az ebédlőben
% látott/tanult adatra black magic jelleg ld 241

\subsection{Global alignment}
\newcommand{\Eset}{\mathcal{E}}
\newcommand{\Iset}{\mathcal{I}}
\newcommand{\Kset}{\mathcal{K}}
\newcommand{\Dset}{\mathcal{D}}

\begin{figure}[H]
    \centering
    \includegraphics[width=\linewidth]{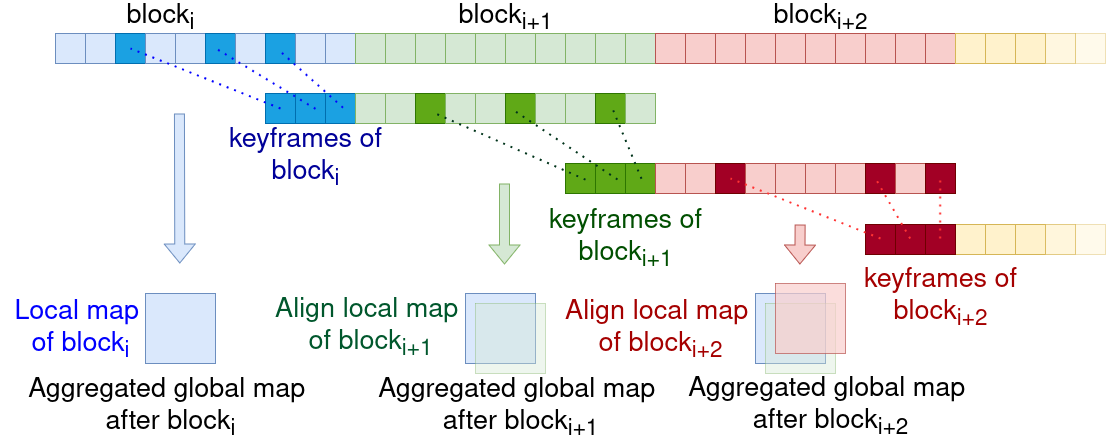}
    \caption{The input stream is partitioned into blocks, with key frames serving as anchors for alignment.}
    \label{fig:align}
\end{figure}

The number of input frames that VGGT can process for 3D generation is constrained by GPU memory. To scale to longer sequences, we construct submaps with VGGT and then align them, thereby making efficient use of memory.
%parameterizes each block’s poses with regard to the block’s first frame. %%
%We partition the input video stream into non-overlapping blocks of consecuting frames.
We partition the input video stream into non-overlapping blocks of consecutive frames.
To align submaps globally and efficiently, we keep a rolling global reference extrinsic matrix and exploit that VGGT 
expresses all poses within in a block of images in the coordinate system of the first frame of the block (see Fig.~\ref{fig:align} for illustration).

%: let \(B_i\) denote the \(i\)-th block.
Let each block \(B_i\) contain \(n\) RGB frames and 
%optionally 
the corresponding depths maps too: 
\[
\Iset^i \coloneqq \{ \mathbf{I}^i_t \}_{t=0}^{n-1},\qquad
\Dset^i \coloneqq \{ \mathbf{D}^i_t \}_{t=0}^{n-1}.
\]

For each block we select \(k\) keyframes (\(0<k<n\)), 
which will provide the basis for alignment:
% OR to provide...: 
\[
\mathcal{K}^i \subset \{0,\dots,n-1\},\quad |\mathcal{K}^i|=k,\qquad
\Kset^i \coloneqq \{ I^i_t \mid t \in \mathcal{K}^i \}.
\]
% We select the keyframes with those of the highest number of detectable features 
The frames with the most detectable features are selected as keyframes 
to suppress motion-blur–induced noise. In the implementation we used the ORB feature detector~\citep{ORB}.
We compose a current frame list $\Iset^{current}$ by concatenating the previous block’s keyframes with the current block’s frames and corresponding depths:
\[
\Iset^{current} \coloneqq \big( (\mathbf{I}^{i-1}_t)_{t\in\mathcal{K}^{\,i-1}},\, (\mathbf{I}^i_t)_{t=0}^{n-1} \big),
\]
\[
\Dset^{current} \coloneqq \big( (\mathbf{D}^{i-1}_t)_{t\in\mathcal{K}^{\,i-1}},\, (\mathbf{D}^i_t)_{t=0}^{n-1} \big).
\]

For $\Iset^{current}$ we predict the depth maps and the extrinsics using VGGT depth and pose heads:
\[
\Dset^{current}_{\mathrm{VGGT}} \coloneqq \{ \mathbf{D}^{current}_{\mathrm{VGGT},j} \}_{j=0}^{n+k-1},
\]
\[
\Eset^{current} \coloneqq \{\mathbf{E}^{current}_j \}_{j=0}^{n+k-1},
\]
where
\[
\mathbf{E}^{current}_j \;=\; \big[\,\mathbf{R}_j \mid \mathbf{t}_j\,\big] \in \mathbb{R}^{3\times 4},
\]
where \textbf{R} denotes rotation and \textbf{t} denotes translation.

The poses within the first block $B_0$ get aligned by VGGT. %by generation. 
For a pair of consecutive blocks $B_{i-1}, B_i$, where $B_{i-1}$ is the last aligned block and $B_i$ is the current, unaligned block, 
we first calculate the depth scale factor and then, the Sim3 similarity transform.%~\citep{Umeyama1991}.

%For appropriate scaling of the estimated depth data we make use of the depth data from the LiDAR sensor. 
Although the depth data of a LiDAR sensor is noisy, and not as fine-grained as VGGT's depth estimate, it can be used 
to obtain a consistent scaling.
%The LiDAR data is noisy, and not as fine detailed as  the VGGT depth estimation has more fine details and we also need to mitigate sensor measurement errors and post-process filtering.
%VGGT depth estimation has more fine details
%We mitigate sensor measurement errors %and post-process filtering.
We apply depth thresholding to the LiDAR depth data to eliminate sensor measurement errors in unreliable ranges (very near and very far) or where the depth estimation confidence of VGGT is too low, by a mask $M_j$, 
that retains only the values between the thresholds for all $D_j \in \Dset^{current}$.
To determine the $s_j$ scaling factor 
we apply least-squares method, as follows:
\[
s_j \;=\; \argmin_{s}\left\|\, M_j\!\big(s\, D^{current}_{\mathrm{VGGT},j} - D^{current}_{,j}\big)\right\|_2^2
\]
for $j=0,\dots,n{+}k{-}1$, where $D^{current}_{\mathrm{VGGT},j}$ 
and $D^{current}$ are the depth maps for frame $j \in [0,n+k-1]$ of the frames in the current frame list, and $M_j$ masks valid pixels, between near and far ranges. 
% megfontolni: a kepen belul is szurni: pl. a lidar filtering hibajabol fakado szelsosegek kompenzalasara. --> journal
We then take the median of the scale values $\tilde{s} = \operatorname{median}\{s_j\}_{j=0}^{n+k-1}$ over the block to improve robustness, and rescale all $D^{current}_{\mathrm{VGGT},j}$ VGGT depth maps in $\Dset^{current}_{\mathrm{VGGT}}$ and the translation component of all extrinsics in $\Eset^{current}$ accordingly:
\[
\mathbf{\hat{D}}^{current}_j \;=\; \tilde{s}\, \mathbf{D}^{current}_{\mathrm{VGGT},j},\qquad
\mathbf{\hat{E}}^{current}_j \;=\; \big[\,\mathbf{R}_j \mid \tilde{s}\,\mathbf{t}_j\,\big].
\]

For the similarity transformation we use our updated extrinsics.
Let us denote the extrinsic matrix of the global graph by $\mathbf{E}_{\mathrm{ref}}$.
As VGGT aligns all camera poses 
%$E_j^{current}, j=1,\dots,n+k-1$ 
in $\Eset^{current}$ with respect to $E_0^{current}$, we pick $E_0^{current}$, to be the corresponding extrinsics for the current frame list:
\[
\mathbf{E}_{ref}^{current} \;=\; \mathbf{E}_0^{current}
\]
%The transformation to align the current block's VGGT poses to all previous, already aligned blocks, is the right-multiplicative alignment
To align the current block's VGGT poses to all previous, already aligned blocks, we apply a right-multiplicative transformation on the extrinsics:
\[
\Delta \;=\; \big(\mathbf{E}_{ref}^{current}\big)^{-1}\,\mathbf{E}_{\mathrm{ref}}.
\]
We update the extrinsics ($\forall E_j \in \Eset^{current}$) in the current lists by this transformation:
\[
\hat{\mathbf{E}}_i \;=\; \mathbf{E}_i\,\Delta,\qquad \forall i,
\]
which aligns the block to the reference while
preserving intra-block relative poses. 
This means, that also the corresponding frame's extrinsic matrix is updated as follows:
\[
\hat{\mathbf{E}}_{ref}^{current} \;=\; \mathbf{E}_{ref}^{current}\,\Delta \;=\; \mathbf{E}_{\mathrm{ref}}.
\]
Therefore, as a last step, we select the key frames from the current block into the keyframe memory
and the global reference extrinsic matrix is set to the (updated) extrinsic matrix of the first keyframe of this current block:
\[
{\mathbf{E}}_{ref} \;=\; \hat{\mathbf{E}}_{ref}^{current}.
\]

To smooth the camera-pose trajectory, we apply the Curvature-Corrected Moving Average (CCMA)~\cite{CCMA}.
For each sample index $n$, CCMA uses the $2k{+}1$-point neighbourhood ${n-k,\ldots,n+k}$ along the trajectory, where $k$ is a hyperparameter; we therefore apply it independently per block.
Because the alignment transform is obtained by composing per-frame transforms, smoothing every transform would compound noise.
We maintain both raw (unsmoothed) and smoothed extrinsics: the smoothed pose is used for submap alignment, while the raw pose is used for point-cloud unprojection. We also compute smoothed extrinsics to produce filtered object point clouds.
This step has three hyperparameters: (1) $k_1$, the number of neighbouring points for curvature correction; (2) $k_2$, the number of neighbouring points for smoothing; and (3) the CCMA kernel. Unless otherwise stated, we keep $k_1$ and $k_2$ at their default values and use a Hann (a.k.a. Hanning) kernel.

%For the first two, we use the default value, although on  sharp turns, it is recommended for the curve correction neighbours to be 1.5--2x less than the smoothing neighbours. The kernel type can be Gaussian or Hanning, out of which we use the latter.
%\textcolor{red}{Curve corrected moving average?}
%\textcolor{red}{Kell a jobb es bal n db szomszéd ahhoz hogy korrigáljon, ezért nem lehet teljesen online per frame csinálni, de blokkonként igen.
%Megvannak az erdeeti exrinsic és a smoothed exrtinsic paramétereket, a smooth-ed újra smootholni akkumulálná a zaj. Csak a kamera pózokat smootholjuk nem az extrinsic-et.
%world to cam extrinsic-et cam to wrold-be, lefuttatjuk a smoothing-ot, ezalapján frissítjük a world cam pose-kat (de ezekkel nem számolunk később a transformation illeszetéshez) illetve visszakonverntáljuk extrinsic-ké és az új blokkra ebből projektálunk pcd-t, ohogy konzisztens legyen.
%3 hiperparam: mennyi mennyi szomszédos vesz jobbra balra figyelembe a curve correctionnél és a smoothingnál, most default-on vannak. Az éles turn-ökhoz a curve correction hyp-nek kisebb kell legyen 1.5---2x mint a másiknak. kernel típus.Gaussian pascal kernel vagy a Hanning, utóbbit használjuk}

\subsection{3D segmentation}
\label{subsec:semanticobj}
\begin{figure}[ht]
    \centering
    \includegraphics[width=0.9\linewidth]{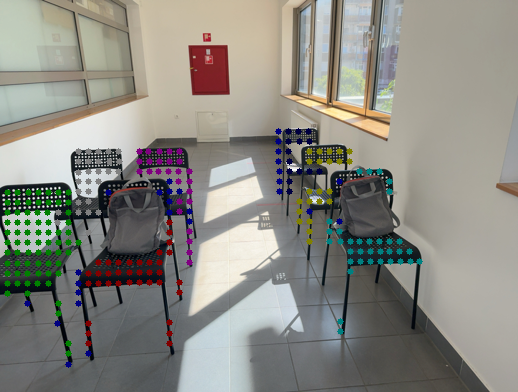}
    \caption{Tracked points sampled from instance masks. Each color corresponds to the assigned instance mask, while dark blue dots indicate unassigned (propagated) points.}
    \label{fig:samplegrid}
\end{figure}

%
%
%\textcolor{red}{Nézd meg, hogy benne van-e, hogy amit a szegmenter ad, az különböző címkéket kap, tehát nem obj, obj, hanem szék, ember, táska... és ezt fiygelembe vesszük, hogy pl. különböző típusokat nem engedünk azonos objektumkét azonosítani.}
%
%
We propose a method to represent real-world objects as 3D point clouds, where the points of each point cloud are labeled with the instance type and ID. 
To ensure correspondence with reality, 
these labels are 
maintained consistently throughout the sequence, thereby defining global objects
in the 3D representation over time.
To add semantic information into the 3D representation 
we apply instance segmentation within the 2D frames, and instance tracking both inside and across blocks.
For 2D segmentation, we use 
a foundation model, namely 
YOLOv9e~\citep{wang2024yolov9} from Ultralytics~\citep{ultralytics_yolo_2023}, which is pretrained on the MS COCO dataset~\citep{mscoco}.
Our goal is to produce per-frame instance masks
that are assigned to a global ID.
We first perform instance segmentation on 2D RGB frames using the foundation model, and post-process the masks with a slight erosion to reduce noise 
after inverse projection back into the 3D representation.
% OR: in the 3D representation after inverse projection.
To exploit the contextual information inherently captured by VGGT,
and ensure that tracking remains robust even in challenging scenarios with numerous overlapping and occluding objects,
we use the tracking head for tracking instance masks across frames.
To improve computational and memory efficiency, instead of tracking all mask pixels, we apply a uniform grid-based subsampling.
The VGGT tracking head returns per-frame point trajectories with a confidence score for all grid points.
We only retain points with confidence scores above a threshold (set to 0.1).
In each frame, each preserved point is assigned to a local instance mask or to the background, yielding an initial association.
Sampling instance masks and the resulting point-to-mask assignments are illustrated in Fig.~\ref{fig:samplegrid}.

% KK:
% 2D instance mask-ok az egyik amivel dolgozunk, per frame
% trackletek: frame-k fölött 
% 
% 
To aggregate 2D instances into 3D objects we perform a voting.
We define a tracklet as a temporal sequence of detections consistently assigned to the same global ID. 
Thus, each tracklet therefore corresponds to a single real-world object across frames. 
%For each frame, we propagate tracked sample points from the previous frame's masks using the VGGT tracking head, and for every current instance mask we count the supporting points: the propagated points that fall inside the mask and belong to a given tracklet (global ID).
The tracked sample points from the masks of the previous frame are propagated to the current frame using the VGGT tracking head, and for every 2D instance mask in the current frame we measure the support from the tracklets, defined as the number of propagated points from a given tracklet (global ID) that fall inside the mask.
%We then perform mutual assignment: for each mask we select the most-supported tracklet, and for each tracklet we select the mask containing the majority of its points.
We then perform mutual assignment: for each mask the tracklet that has the largest support on that mask is selected as a candidate, and then each tracklet is assigned to the mask (out of the ones where it became a candidate) on which most of its points fall.
This conservative strategy (favoring false negatives over false positives) mitigates occlusion errors and avoids erroneous merges. We also enforce a type-consistency constraint: a mask and tracklet are matched only if their object classes agree (e.g., a bag may not become a chair).
%If a mask has no valid match in the voting/support matrix, we allocate a new global ID and initialize a new tracklet; 
If a mask has no valid mutual match among the tracklets, then a new tracklet is initialized and a new global ID is allocated to it if it happens on the first frame of a block, otherwise a default ID ("untracked") is assigned to it; 
if a tracklet receives no supporting points in a frame, we record an empty 
state to preserve temporal alignment and allow for later re-identification.
Unmatched tracklets may get resolved via re-identification.

For re-identification across temporal gaps or block boundaries for the very same objects, the point cloud of all newly initialized tracklet is compared against the point clouds of previously established tracklet. 
A cost matrix is constructed using the 3D intersection-over-union (IoU) of bounding boxes of the point clouds. 
If the overlap exceeds a predefined threshold, the new sub-tracklet is merged with the existing tracklet, and all stored identifiers are updated accordingly, both at the instance level and at the global object level.

Objects may leave and re-enter the field of view, so we maintain a persistent memory of previously seen instances. 
When a new tracklet appears, we perform re-identification to check whether it corresponds to a past object. 
For each 2D instance, we compute the median point of its 3D point cloud (PCD) for every frame in the block—yielding up to one median point per frame—and label these with the global ID. 
We do the same for the current block’s instances that could start new tracklets. 
We then compare historical and current median-point clouds using Chamfer distance; if the closest match is below a threshold (a hyperparameter, 30 cm in our setup), we match the instance to the prior tracklet.
This approach also handles the type of noise occurring when 2D instance masks cover holes, which would otherwise appear as artifacts during 3D projection.

% \textcolor{red}{
% Mivel vannak objektumok amik idorol idore kikerülnek a field of view-bol aztan pedig vissza, ezert a memoriankban, mint emberi memoria, nem vram, taroljuk a modellben az eddig latott objektumokat. Ha jon egy uj tracklet, akkor megnezzuk hogy nem lehet-e re-identifikalni vele egy korbbit, azaz nem bukkant-e fel egy mar latott objektum ujra. 
% A regi objektumokra, azaz amit eddig valaha is lattam, akar most is, de lehet hogy most epp ebben a blokkban nem, de benn vannak a 3D reprezentciomban (pcd) kiszamoljuk a pcd median-t. instance-enkent 1-1 medianpont, azaz blokkonkent 10 db pont ha 10 frame van a blokkban, és a global ID címkézi kvázi pjet.
% A mostani instance-kre amikkel uj trackletet inditanek is kiszamolom (ez max 10 db medianpont).
% Megnezem, hogy a mostaniak es a korabbiakra kapott median felhőkra a Chamfer distance-ek alapjan melyik van a legkozelebb, es egy adott thresh alatt (hyperparam, most epp 40 cm nálunk) van, akkor osszerendeljuk.  
% }

\subsection{Change detection}
Apart from the static elements of the environment, there are realistically changing parts. 
We target to handle changes in the state or presence of objects --- for example, a chair transitioning from unoccupied to occupied, or vice versa --- where the actual movement does not occur within the camera’s field of view.
The proposed pipeline detects changes by assigning timestamps to global objects, providing a memory mechanism through continuous updating.

Frames arrive sequentially, thus we assign incrementally an index to each frame, and each segmented instance is assigned the index of the frame in which it appears.
This enables us to track the temporal occurrence of each instance. 
We associate the instance with its corresponding global ID (i.e., the object it represents) as described in subsection~\ref{subsec:semanticobj}, which allows us to maintain a memory of the object and its state changes over time.
%When the current instance mask of an object deviates from the expected state,
%If detections in the current frame indicate that the object is no longer present, we update its memory state as follows.
%We update memory only for objects whose latest index is not from the last block.
Each object can be in one of three states: (1) \textsc{Recent}, (2) \textsc{Retained}, or (3) \textsc{Removed}.  
We associate each object with a confidence value $c$.
An object is considered \textsc{Recent} if it is observed in the last block, and its confidence is set to $c=1$.

The states of the objects are updated at each block.
The main steps are described in Alg.~\ref{alg:memory}.
To update the state of each object $o$ with 3D points $P_o$, in the subsequent frames, we check the visibility.
We project all points onto the 2D image plane of the current camera pose $\mathbf{T}_{ts}$, where $ts$ is the timestamp (or index) of the frame, yielding the projected image $\hat{I}_{i}$. 
We keep only those points with depth $z_{\text{proj}} > 0$ (i.e., in front of the camera plain) and with pixel coordinates inside the image bounds; this subset is denoted by $\Omega_i^o$.
If $|\Omega_i^o|=0$ the object is not in the field of view, meaning it is not observed in the latest block, and becomes \textsc{Retained}.
On this state change no penalty is applied and the last value of $c$ is maintained, which may then be gradually decayed over time.

For objects that are within the field of view, we further evaluate their visibility using the depth image $\mathbf{D}_i$. 
A projected pixel $p \in \Omega^i_o$ is considered \textit{visible} if
\[
z_{\text{proj}}(p) \;\leq\; z_{\text{obs}}(p) + \delta,
\]
where $z_{\text{proj}}(p)$ is the depth of the projected 3D point, $z_{\text{obs}}(p)$ is the observed depth at the corresponding pixel location, and $\delta$ is a tolerance value (e.g. $1mm$). This formalizes the intuition that if no other objects are present between the expected object and the camera, the object’s mask should be visible in the image. If the mask is not observed under these conditions, the object is considered missing.

The fraction of visible pixels is then defined as
\[
f_{\text{vis}} \;=\; \frac{\# \text{ visible pixels in } \Omega_o}{|\Omega_o|}.
\]

If $f_{\text{vis}} \geq \tau_{\text{vis}} $ threshold value for visibility ($\tau_{\text{vis}}=0$ by default) but no detection is associated with object $o$ in the current frame, the memory confidence is decayed as
\[
c \;:=\; c - \eta , 
\qquad \eta,c \in [0,1].
\]

The parameter $\eta$ reflects the confidence in forgetting: smaller values make the system less certain about forgetting (slower decay), while values closer to 1 increase the confidence and lead to faster forgetting.
If the confidence $c$ of an object becomes zero, then its state is set to \textsc{Removed}.

To avoid penalizing small boundary fragments that detectors frequently miss, decay is skipped when the projected area fraction falls below a threshold ($\tau_{\text{area}}=0$ by default). 
%Depth-based occlusions are naturally accounted for, as they reduce $f_{\text{vis}}$ and thereby prevent decay (e.g., when an object is partially hidden behind a seat backrest).
% For object within the field of view we use the depth image $\mathbf{D}_i$, a pixel $p \in \Omega_o$ is ``visible'' if
% \[
% z_{\text{proj}}(p) \leq z_{\text{img}}(p) + \delta.
% \]
% The visible fraction is
% \[
% f_{\text{vis}} = \frac{\# \text{ visible pixels in } \Omega_o}{|\Omega_o|}.
% \]
%
% If $f_{\text{vis}} \geq \tau_{\text{vis}}$ but no detection is assigned to $o$ in this frame, decay its memory confidence:
% \[
% c := \eta \cdot c, \quad \eta \in (0,1);
% \]
% otherwise leave $c$ unchanged. To avoid penalizing tiny slivers near the image edge that detectors often miss, skip decay when the projected area fraction is below $\tau_{\text{area}}$ (default $0$). Depth-based occlusions naturally reduce $f_{\text{vis}}$, preventing decay (e.g., a seat backrest in front of the object).
Objects with a \textsc{Retained} or \textsc{Removed} status may later become active again via re-identification.

% \textcolor{red}{
% if detections in the current frames indicate that the object is no longer present in the scene, the representation is likewise updated, as described below....
% azok az objektumokra amik a temp index nem az utolsó blokkból van (mert ami onnan van, azt biztos látjuk), azokra megnézzük a field of view és mélységkép alapján hogy az adott kamera pózból látnunk kell-e az obejktumot, és ha nem egyezik, akkor elvault az objektum jelenlétéről számontartott állapot, ezért discardoljuk ha nincs ott.}
% %We apply confidence-based aggregation for stability. 
% % todo: le kellene írni az alg lépéseit mondatosan is
% The main steps are described in Alg.~\ref{alg:memory}.
% \textcolor{red}{három állapota van egy obejtumnak: emlék, vagy discard, vagy friss. Friss ami a legutolsó blokkból jött, ha nem, akkor emlékké válik. Az emlékekhez (minden obj-hoz) vagy egy memconfidence, a friss = 1, ha emlék akkor nem változik  de! ha egy blokkon belül látnom kéne de nem látom akkor rate mentén csökkentem. Ha megint kikerül ez a rész, akkor ott marad ameddig csökkent. Háttámla miatt a depth nem decay-eli. Thresh: mekkora \% kell az objektumnak hogy "látnom kéne" --> ha túl pici az ibj széle, hogy a YOLO felszedje, ezt szűrjük, most 0.}

%\newcommand{\myalgorithm}{%
%\begingroup
%\removelatexerror% Nullify \@latex@error
\begin{algorithm}%[H]
\caption{Object Visibility and Confidence Update. We update object visibility states as the last step of processing the current block of frames.}\label{alg:memory}
\KwIn{Sequence of frames with object detections}
\KwOut{Updated object states with confidence values}
\ForEach{object $o$}{
    Reproject $o$ into current block (using current poses)\;
    \If{$o$ is visible in the reprojected view}{
        Compare depth of $o$ with scene depth\;
        \If{scene depth $<$ object depth}{
            Increase uncertainty value of $o$ as it may be occluded\;
        }
    }
    Determine state of $o$: in-field-of-view / hidden\;
    \If{$o$ not present where expected}{
        Update confidence of $o$ (simulate forgetting)\;
    }
}
\end{algorithm}
%\endgroup}
% Steps:
% 1. Iterate over all types of objects: reproject the 3D objects to check if it is visible in the current block -- ie. from the current few poses. 
% 2. is it blocked: based on the depth: if the depth from the pose is smaller (ie. something is closer) than the object, then it may be occluded: increase uncertainty value
% % --> mi van ha messzebb van az első? annak is kéne befolyásolni! 
% 3. annak függvényében, hogy ott van-e ahol a visszavetítés alapján kéne lennie vagy in-field-of-view vagy hidden
% 4. frissítés: ha nincs ott, akkor a konfidenciát állítjuk, kvázi felejtünk.

\subsection{Ego-location and object positions}
In the streaming scenario, we consider the position of the user as the camera pose corresponding to the last frame index.
To determine the distances of detected objects from the current pose (i.e. the user holding the sensor) and the distances among objects we use a matching approach as follows.
%Whenever we update an object, we also update its point cloud. We compute a median over the point cloud.
%This is a per-object median rather than a per-instance median as in Section~\ref{subsec:semanticobj}.
%We regard this median as the object’s location.
Whenever an object is updated, its associated point cloud is updated as well. A median point is then computed over this point cloud, considering all points belonging to the object (unlike the per-instance median used in Section~\ref{subsec:semanticobj}). This median serves then as the object’s estimated location.

In each iteration, i.e. in each block, we compute the relative positions for all objects currently considered valid, using the position of the user and the object’s centroid obtained from the median, using Euclidean distance.
% \textcolor{red}{
% Azt vesszük a saját pozíciónknak, ami az utolsó frame index-hez tartozó world-to-camera pose.
% Minden objektum, amikor frissítjük: update-eljük a point cloudjait.
% A pointcloudra számolunk egy mediánt. 
% Note. ez nem ugyanaz teljesen mint az előző fejezetecskében: mert ez itt már per-object median, és nem per instance median.
% Ezt tekintjük a lokációjának.
% Minden iterációban, azaz minden blokkban kiszámoljuk az összes jelenleg validnak gondolt objektumra a relatív pozíciókat a mi pozíciónktól és az obj mediánnal kapott középpontjától.
% Euklideszi távolságot néz.
% }

% Ábra: https://app.diagrams.net/#G1HTlBrEszw_uPqVjpmzrLL1c8pp1w5PW7

%% file: vggt/4_results.tex
\section{Results}
\label{sec:result}

%% 
% To benchmark the proposed method according to the task complexity of finding an empty seat, we consider three levels.
% The baseline is defined by the Cybathlon rules, meaning there are exactly six chairs in a $2\times3$ layout, two persons, two bags, and everything is considered as static.
% A more challenging case assumes an unknown number of objects and an unknown chair layout. Finally, a more general case allows objects (persons or bags) to change position while out of view, requiring detection of such changes. 

\subsection{Experimental setup and metrics}
For evaluation, we use the TUM RGB-D dataset~\citep{sturm2012tum}, a standard benchmark for indoor perception. It provides calibrated, time-synchronized RGB-D sequences across diverse layouts with accurate motion ground truth, making it well suited for assessing visual odometry, SLAM, and map building—the backbone of navigation stacks.
We also evaluate on the 7-Scenes dataset~\citep{sevenscenes}, which includes dense point clouds, allowing us to measure accuracy, completion, and Chamfer distance, mirroring the VGGT-SLAM protocol. 
We report average values in Table~\ref{tab:ssc} measured on sequences: \emph{stairs, pumpkin, heads}.
For navigation and SLAM we report Absolute Trajectory Error (ATE)~\citep{sturm2012tum}.
We evaluate odometry using the \emph{evo} toolkit~\citep{grupp2017evo}.

For semantics and 3D reconstruction, we target a use case relevant to visually impaired users: identifying unoccupied seats in public indoor settings. As, to our knowledge, no widely adopted benchmark exists for this task, we provide qualitative results on our own dataset. 
We present visualizations of reconstructed 3D maps --- both RGB and semantic --- and include scenarios with environmental changes (e.g., a bag displaced from a chair).

Measurements were performed on a computer with Ubuntu 22.04, equipped with an 
AMD Ryzen Threadripper Pro 5955WX CPU, NVIDIA RTX 4090 GPU with 24GB of VRAM, and 512GB of RAM. 

Similarly to the evaluation protocol used in VGGT-SLAM~\citep{vggtslam}, we apply a transformation to align the predictions with the ground-truth coordinate system, enabling direct comparison.

\subsection{Quantitative results}
VGGT can process about 60 frames within the 24 GB VRAM limit.
For the proposed pipeline we measured VRAM usage with NVIDIA's Nsight Systems tool~\citep{nvidia_nsight_systems}.
Although the implementation has not been optimized, we report the measurements in Table~\ref{tab:tumtimeetal}., alongside the total running time and the used CPU RAM as a reference, as well as the number of frames of the individual scenes.

% \onecolumn
% \begin{table*}[!htbp]
% \centering
% \caption{RAM, VRAM, and running time measurement on the 7-scenes dataset}
% \label{tab:sctimeetal}
% \begin{tabular}{lccccccc}
% \toprule
% Sequence & chess &fire &heads &office &pumpkin &kitchen & stairs \\
% %\midrule
% Number of frames  &  &  &  & &  &  &   \\
% \midrule
% Max RAM              &      &      &      &      &      &      &      \\
% Max VRAM              &      &      &      &      &      &      &      \\
% Total time            &      &      &      &      &      &      &       \\
% \bottomrule
% \end{tabular}
% \end{table*}

\begin{table*}[htbp]
\centering
\caption{RAM, VRAM, and running time measurement on the TUM RGBD dataset}
\label{tab:tumtimeetal}
\begin{tabular}{lccccccccc}
\toprule
Sequence & 360& desk &desk2 &floor &plant &room& rpy &teddy &xyz  \\

\midrule
Number of frames  & 740 & 570 & 620 & 1220 & 1120 & 1350 & 690 & 1400 &  790 \\
\midrule
Max RAM (GB)  &   12.4   &   10.1   &  10.8 & 18.7 &  17.3 &    20.3 & 11.6 & 21.0  & 13.0 \\
Max VRAM (GB)  &   17.8   &   17.8   &   17.8  & 17.7 & 17.8 & 17.8 & 17.8 & 17.8 & 17.9 \\
Total time (s) &    134.53 &    110.09   & 115.61 &  211.66 & 193.63 & 231.14  & 134.57  & 247.91 &   147.51   \\
\bottomrule
\end{tabular}
\end{table*}

%\subsection{Trajectory}
In Tables~\ref{tab:sevscenesSLAM} and
~\ref{tab:TUMSLAM} we report the absolute trajectory error (ATE) as root mean square error (RMSE) for the 7-Scenes and for the TUM RGB-D sequences, respectively.
% From the VGGT SLAM paper:
%"One particular scene where our method underperforms in on the TUM floor scene. This highlights a challenge of estimating homography, which is the presence of degeneracy in the case of a planar scene. The floor scene contains several images that only view the flat floor leading to non-unique solutions for the homography matrix, which causes the overall reconstruction to diverge. Building robustness for the planar case is an important component for SL(4) SLAM, which we leave as an exciting direction for future work."
Our approach works remarkably well on the TUM floor scene, which proved to be a very challenging subtask for VGGT SLAM, because, as they note, the high number of images containing only the flat floor lead to non-unique solutions for the homography matrix, and ultimately divergence. By using shorter block sizes and thereby limiting the focus of the VGGT reconstruction to the more recent frames, we successfully achieved much better convergence.

%%%%%%%%%%%%%%%%%%%%%%%%%%%%%%%%%%%%%%%%%%%%%%%%%
%%           SLAM OSSZEVETES 7-scenes adatra
%%%%%%%%%%%%%%%%%%%%%%%%%%%%%%%%%%%%%%%%%%%%%%%%%
\begin{table*}[!htbp]
\centering
\caption{Comparison of the root mean square error of VGGT-SLAM and the proposed method across 7-Scenes sequences.}
\label{tab:sevscenesSLAM}
\begin{tabular}{lcccccccc}
\toprule
Method & chess &fire &heads &office &pumpkin &kitchen & stairs & Avg \\
\midrule
VGGT-SLAM Sim(3)  &   0.037  & 0.026  & 0.018  & 0.104  & 0.133  & 0.061 & 0.093  & 0.067     \\
VGGT-SLAM SL(4)   &   0.036  & 0.028  & 0.018  & 0.103  & 0.133  & 0.058  & 0.093  & 0.067      \\
Ours            &   0.055   &   0.047   &  0.034   &  0.120   &   0.152   & 0.063    &       0.032    & 0.072     \\
\bottomrule
\end{tabular}
\end{table*}
\noindent

\begin{table*}[htbp]
\centering
\caption{Comparison of the root mean square error of VGGT-SLAM and the proposed method across TUM RGB-D sequences.}
\label{tab:TUMSLAM}
\begin{tabular}{lcccccccccc}
\toprule
Method & 360& desk &desk2 &floor &plant &room& rpy &teddy &xyz & Avg \\
\midrule
VGGT-SLAM Sim(3)  &   0.123  & \cellcolor{green!10}0.040  & 0.055  &  0.254  & \cellcolor{green!40}0.022  & \cellcolor{green!10}0.088  & 0.041  & \cellcolor{green!40}0.032  & \cellcolor{green!10}0.016  & 0.074  \\
VGGT-SLAM SL(4)   & \cellcolor{green!40}0.071  &  \cellcolor{green!40}0.025   & \cellcolor{green!10}0.040   & \cellcolor{green!10}0.141   & \cellcolor{green!10}0.023  &  0.102   & \cellcolor{green!10}0.030   & \cellcolor{green!10}0.034  &  \cellcolor{green!40}0.014   & \cellcolor{green!40}0.053 \\
{Ours} & \cellcolor{green!10}{0.118} & {0.042} & \cellcolor{green!40}{0.039} & \cellcolor{green!40}{0.063} & {0.077} & \cellcolor{green!40}{0.081} & \cellcolor{green!40}{0.029} & {0.076} & {0.034} & \cellcolor{green!10}{0.062} \\
\bottomrule
\end{tabular}
\end{table*}

\noindent

%%%%%%%%%%%%%%%%%%%%%%%%%%%%%%%%%%%%%%%%%%%%%%%%%
%%           POINT CLOUD OSSZEVETES
%%%%%%%%%%%%%%%%%%%%%%%%%%%%%%%%%%%%%%%%%%%%%%%%%

\begin{table*}[!htbp]
\centering
\caption{Root mean square error (RMSE) of the reconstruction evaluated on 7-Scenes}
\label{tab:vggt_slam_7scenes}
\begin{tabular}{lccc}
\toprule
Method & Acc. $\downarrow$ & Complet. $\downarrow$ & Chamfer $\downarrow$ \\
\midrule
VGGT-SLAM Sim(3) &0.052 &0.062 &0.057 \\
VGGT-SLAM SL(4)  &0.052 &0.058& 0.055  \\
Ours  &     0.0566   &  0.0668  &  0.06179     \\
\bottomrule
\label{tab:ssc}
\end{tabular}
\end{table*}
%\twocolumn

\subsection{Qualitative results}
Beyond the quantitative results, the method’s potential is well illustrated by how it performs on difficult cases. In Fig.~\ref{fig:vggt_vs_ours_2x4}, we show the point clouds produced by VGGT-SLAM and by our method for the same input.

We also provide illustrative videos\footnote{
\href{http://users.itk.ppke.hu/~horan3/SceneVGGT/SceneVGGTalign.mp4}{
Alignment video}}$^{,}$\footnote{\href{http://users.itk.ppke.hu/~horan3/SceneVGGT/SceneVGGTsemantics.mp4}{Semantic overlay video}}. 
Firstly, of the alignment process, showing how the 3D map is built up iteratively. The visualization is recorded in real time without any speedup, demonstrating the actual performance of the pipeline (including the rendering). 
Secondly, a video that presents the same sequence with semantic labels overlaid on the reconstructed 3D representation.
%\textcolor{red}{We provide additional point cloud images in the Supplementary material.}

\captionsetup{skip=4pt}
\captionsetup[sub]{font=small,justification=centering}
\newcommand{\imgH}{30mm} 
\begin{figure*}[htbp]
  \centering

  % ===== Row 1: VGGT-SLAM =====
  \textbf{VGGT\textendash SLAM}\par\vspace{0.35em}

  \makebox[\textwidth][c]{%
    \hspace*{\fill}%
    \begin{subfigure}[t]{0.32\textwidth}\centering
      \includegraphics[height=\imgH,keepaspectratio]{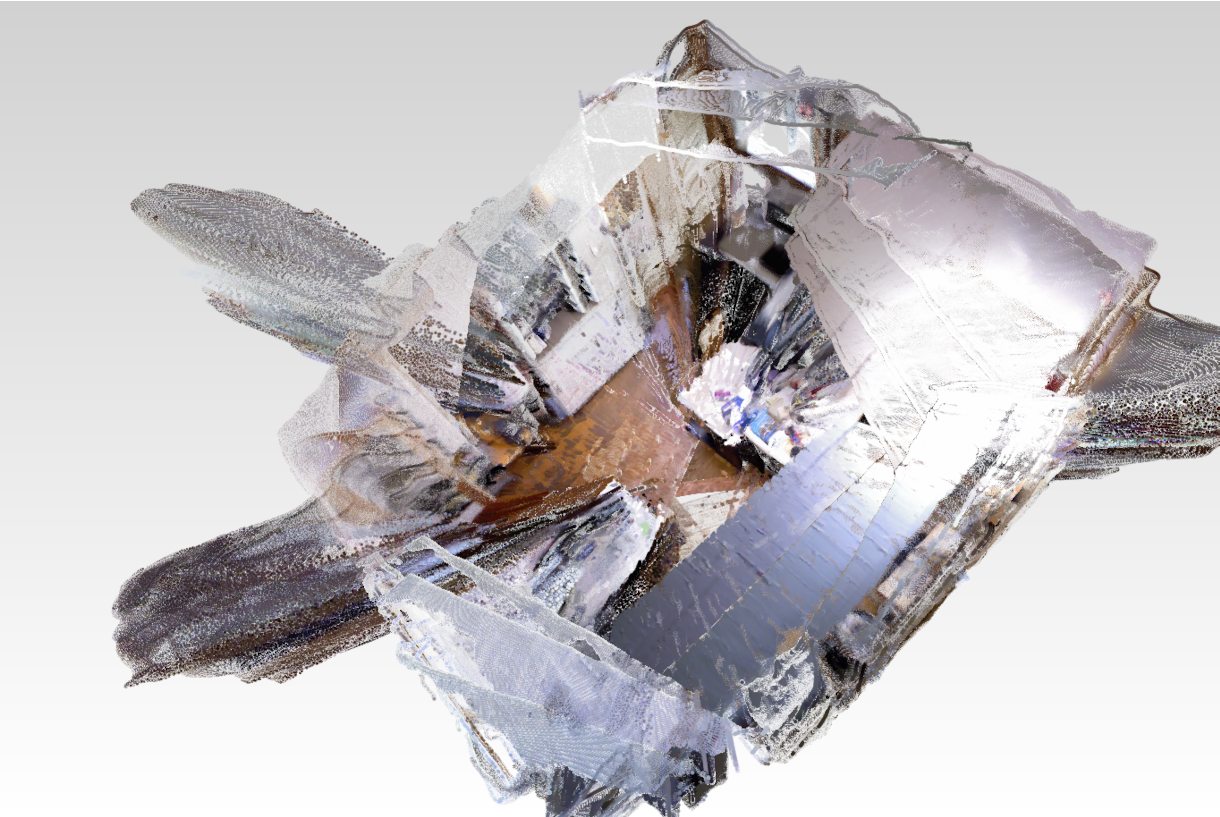}
      \caption{TUM RGB-D 360}
      \label{fig:vggt_01}
    \end{subfigure}%
    \hspace{\fill}%
    \begin{subfigure}[t]{0.32\textwidth}\centering
      \includegraphics[height=\imgH,keepaspectratio]{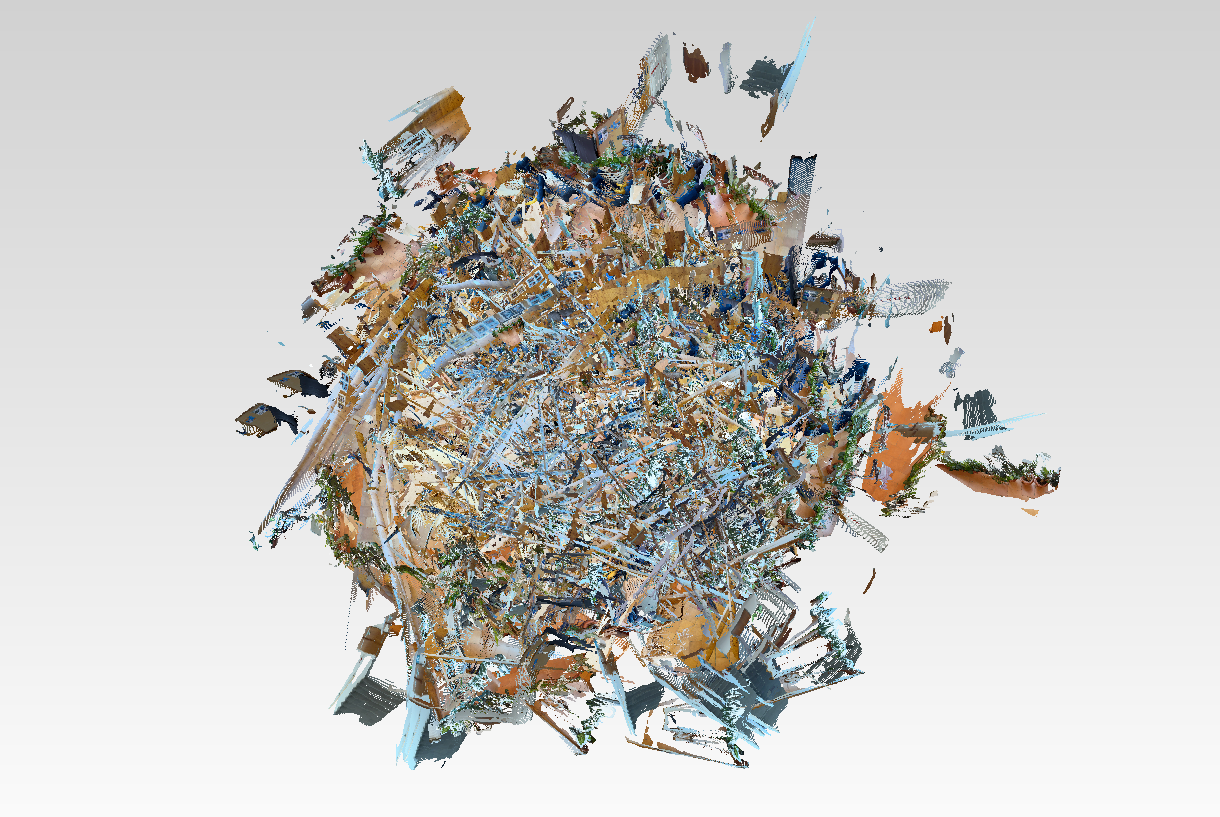}
      \caption{Custom data seq.\#1 (large hall)}
      \label{fig:vggt_02}
    \end{subfigure}%
    \hspace{\fill}%
    \begin{subfigure}[t]{0.32\textwidth}\centering
      \includegraphics[height=\imgH,keepaspectratio]{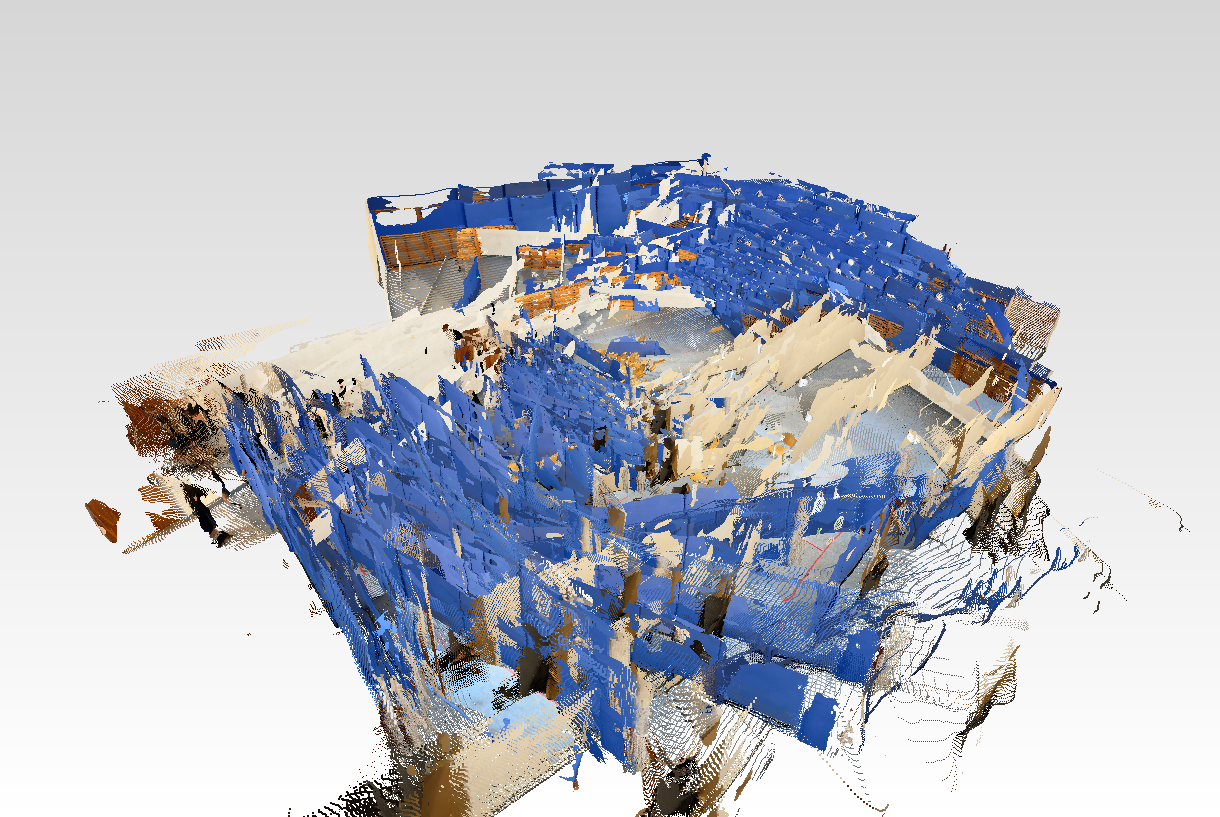}
      \caption{Custom data seq.\#2 (large hall)}
      \label{fig:vggt_03}
    \end{subfigure}%
    \hspace*{\fill}%
  }

  \vspace{0.9em}

  % ===== Row 2: Ours =====
  \textbf{Ours}\par\vspace{0.35\baselineskip}

  \makebox[\textwidth][c]{%
    \hspace*{\fill}%
    \begin{subfigure}[t]{0.32\textwidth}\centering
      \includegraphics[height=\imgH,keepaspectratio]{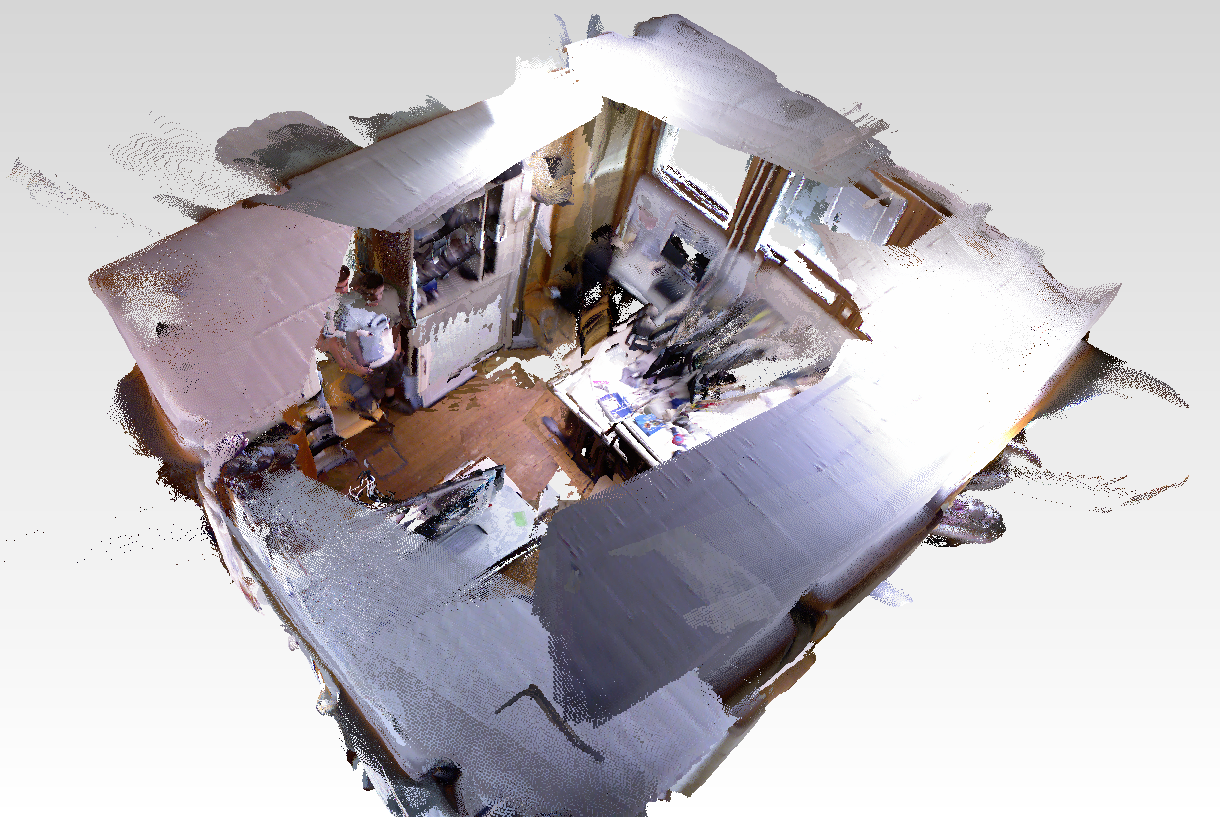}
      \caption{TUM RGB-D 360}
      \label{fig:ours_01}
    \end{subfigure}%
    \hspace{\fill}%
    \begin{subfigure}[t]{0.32\textwidth}\centering
      \includegraphics[height=\imgH,keepaspectratio]{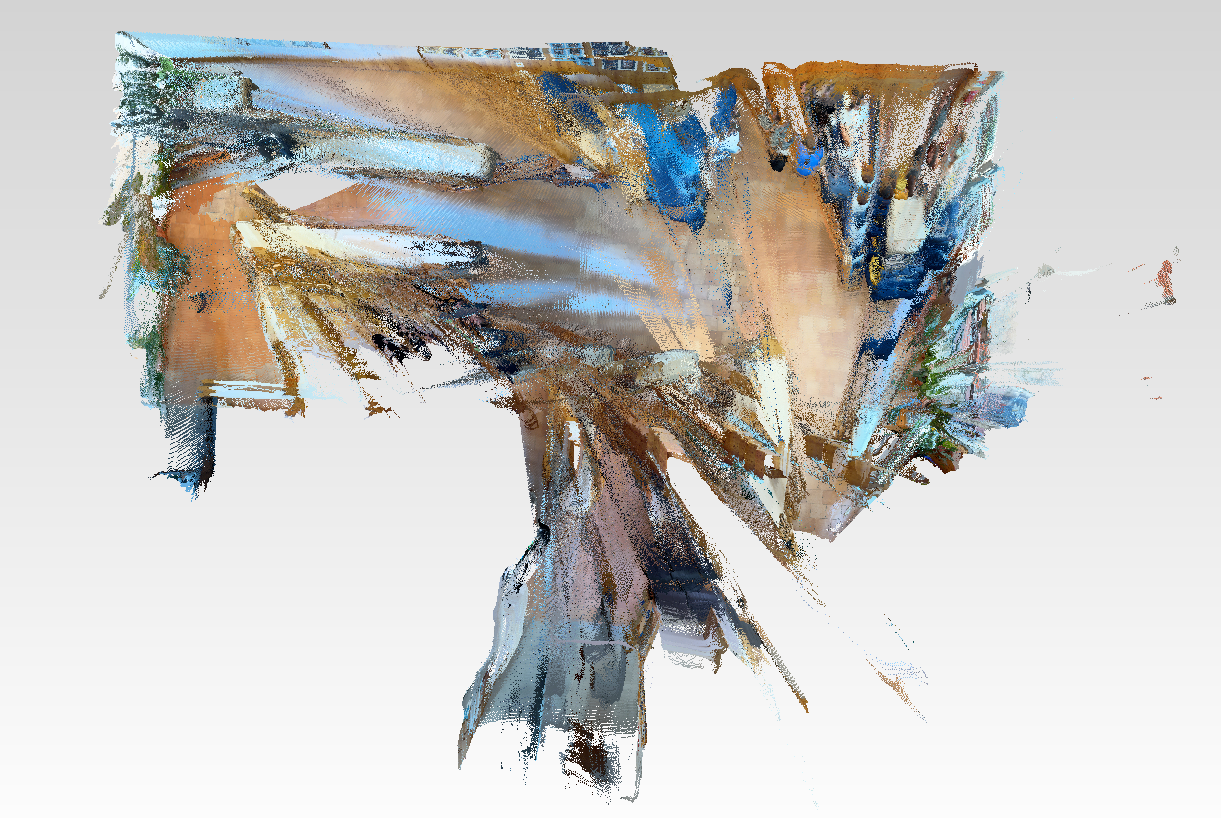}
      \caption{Custom data seq.\#1 (large hall)}
      \label{fig:ours_02}
    \end{subfigure}%
    \hspace{\fill}%
    \begin{subfigure}[t]{0.32\textwidth}\centering
      \includegraphics[height=\imgH,keepaspectratio]{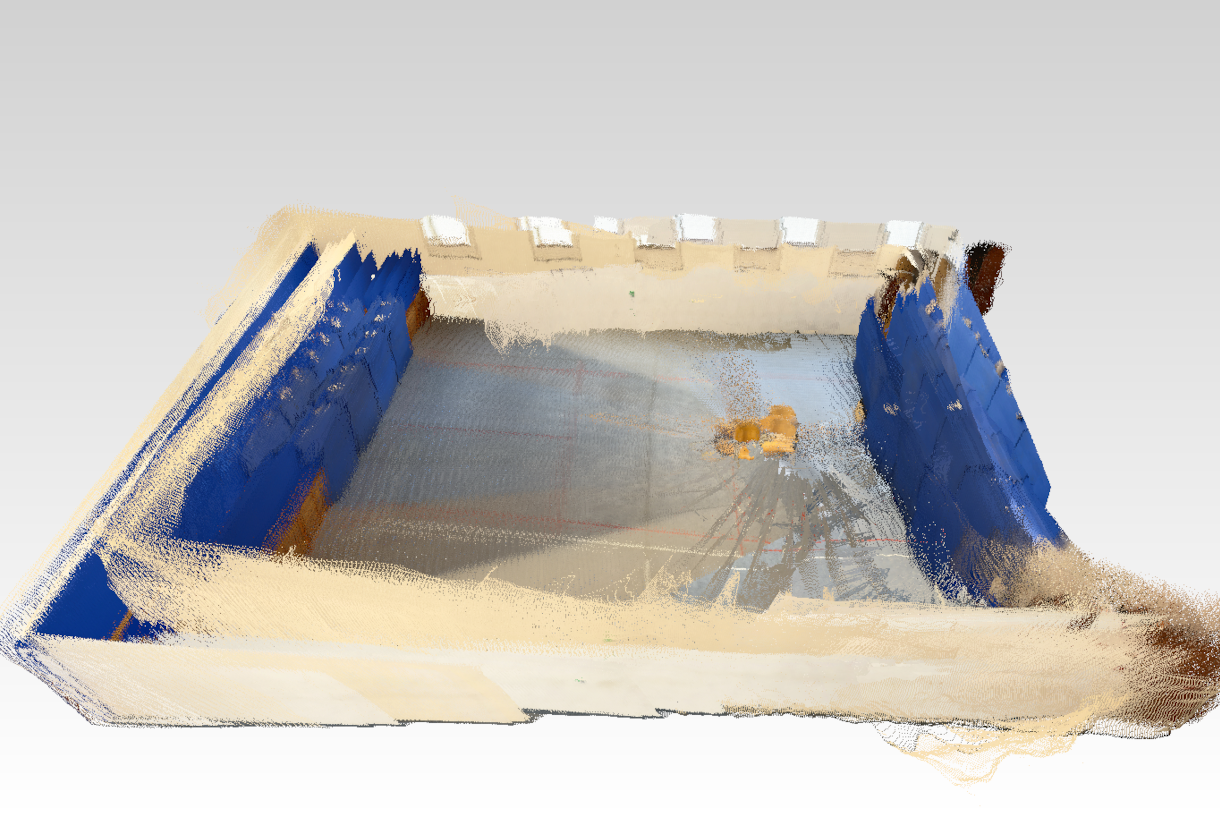}
      \caption{Custom data seq.\#2 (large hall)}
      \label{fig:ours_03}
    \end{subfigure}%
    \hspace*{\fill}%
  }

  \caption{Point-cloud comparison on identical inputs across four sequences.
  Top row: VGGT\textendash SLAM; bottom row: our method. Vertically aligned pairs correspond to the same input.}
  \label{fig:vggt_vs_ours_2x4}
\end{figure*}

% todo: segmentation results.

%% file: vggt/5_conclusion.tex
\subsection{Discussion}

In instance tracking, when an instance has no tracklet based on the tracked points, we only initialize a tracklet if this occurs at the very first frame in the frame list. 
The reason is that the VGGT tracking head is initialized from the first frame of the block. 
This means that for intermediate frames within the block (e.g., when an object is occluded), no new tracklet is created that could later be re-identified. 
Instead, such objects lose or do not get a global object ID and remain untracked but detected objects, if they are not visible at the beginning of the block.
A possible way to address this would be to wait for data within each block and decide, using some method, which frame in the block should be used for initializing tracking, or to allow initialization from multiple frames. However, the latter approach has the drawback that each initialization and tracking step takes about $0.5$ seconds, which would introduce too much overhead. 
The former also incurs computational overhead and, on its own, may not eliminate all failure cases: with a block size of $10$, an object can remain untracked for about $0.5$s in the worst case and thus becomes “untracked” --- we know it is present but cannot associate it with a specific instance --- which has a negligible impact on the assistive navigation application.

\subsection{Conclusion}

We introduced a memory-efficient pipeline for temporally coherent 3D semantic mapping based on Vision Gated Generative Transformers (VGGT).
By processing video input in blocks and aligning submaps, the method overcomes the high memory footprint of standard VGGT, enabling long-sequence operation.
Our integration of semantic instance masks with VGGT’s tracking head yields object-level consistency across time, supporting robust change detection in dynamic environments. 
With the representation now containing all necessary information, the remaining step is to communicate it effectively to the user for assistive navigation.
The framework is designed to enable the processing streaming input and achieves close to real-time performance on a desktop GPU, making it well suited for assistive navigation and other applications where both geometric accuracy and semantic awareness are required. 
Future work will explore the integration of moving, dynamic objects in the representation.

\section{Acknowledgments}
\label{sec:acknow}

The work was supported by the TKP2021-NVA-27 grant, funded by the Ministry of Innovation and Technology with
support from the National Research Development and Innovation Fund under the TKP2021 program.
Gergely Dinya was supported by the Research Fellowship EKÖP-25-2-I-ELTE-987 of the Ministry of Culture and Innovation of Hungary, funded by the National Fund for Research, Development and Innovation (NRDI).